\documentclass[letterpaper]{article} 
\usepackage{aaai24}  
\usepackage{times}  
\usepackage{helvet}  
\usepackage{courier}  
\usepackage[hyphens]{url}  
\usepackage{graphicx} 
\usepackage{natbib}  
\usepackage{caption} 
\DeclareCaptionStyle{ruled}{labelfont=normalfont,labelsep=colon,strut=off} 

\usepackage{epstopdf}
\usepackage{paralist}
\usepackage[linesnumbered,ruled,vlined]{algorithm2e}
\usepackage{amssymb}
\usepackage{amsmath}
\usepackage{etoolbox}
\usepackage{tikz,pgfplots}
\usepackage[export]{adjustbox}
\usepackage{subcaption}
\usepackage{tabulary}
\usepackage{booktabs} 

\usepackage{multirow}
\usepackage{todonotes}

\newcommand{\MONTH}{%
	\ifcase\month
	\or January
	\or February
	\or March
	\or April
	\or May
	\or June
	\or July
	\or August
	\or September
	\or October
	\or November
	\or December
	\fi}

\newcommand{\bv}[2]{\mathbf{#1}_\mathrm{#2}} 

\newcommand{\OBST}{\mathcal{O}}

\newcommand{\model}{\mathcal{M}}
\newcommand{\Dataset}{\mathcal{D}}

\newcommand{\node}{n}
\newcommand{\nodeI}{n_{\mathrm{I}}}
\newcommand{\nodeG}{n_{\mathrm{G}}}
\newcommand{\OPEN}{\textsc{Open}}
\newcommand{\CLOSED}{\textsc{Closed}}	

\newcommand{\childnodes}{\bv{n}{}'}
\newcommand{\childcolnodes}{\bv{n}{\mathrm{C}}'}

\newcommand{\greedyEuc}{h_{\textsc{EUC}}}

\begin{document}

\nocopyright{}

\title{SLOPE: Search with Learned Optimal Pruning-based Expansion}




\author{Davor Bokan\textsuperscript{\rm 1},
Zlatan Ajanovi\'{c}\textsuperscript{\rm 2},
Bakir Lacevic\textsuperscript{\rm 1}
}

\affiliations{
    \textsuperscript{\rm 1} University of Sarajevo\\
    \textsuperscript{\rm 2}RWTH Aachen University


    \textrm{dbokan1@etf.unsa.ba, zlatan.ajanovic@ml.rwth-aachen.de, bakir.lacevic@etf.unsa.ba }

}

\maketitle

\begin{abstract}
Heuristic search is often used for motion planning and pathfinding problems, for finding the shortest path in a graph while also promising completeness and optimal efficiency. The drawback is it's space complexity, specifically storing all expanded child nodes in memory and sorting large lists of active nodes, which can be a problem in real-time scenarios with limited on-board computation. To combat this, we present the Search with Learned Optimal Pruning-based Expansion (SLOPE), which, learns the distance of a node from a possible optimal path, unlike other approaches that learn a cost-to-go value. The unfavored nodes are then pruned according to the said distance, which in turn reduces the size of the open list. This ensures that the search explores only the region close to optimal paths while lowering memory and computational costs. Unlike traditional learning methods, our approach is orthogonal to estimating cost-to-go heuristics, offering a complementary strategy for improving search efficiency. We demonstrate the effectiveness of our approach evaluating it as a standalone search method and in conjunction with learned heuristic functions, achieving comparable-or-better node expansion metrics, while lowering the number of child nodes in the open list. Our code is available at https://github.com/dbokan1/SLOPE.

\end{abstract}


\section{Introduction}

The problem of path planning is a well-known and researched topic in the fields of robotics and artificial intelligence. Oftentimes, we look at a graph search task formulation with the objective of finding the shortest path to the goal with the least amount of visited nodes. To solve the problem we deploy search algorithms such as A* \citep{Hart1968AFB} or best-first search, which use a heuristic to guide node expansions until reaching said goal. Classic heuristics, such as Euclidean or Manhattan distance, are very simple and thus do not provide great results on more complicated tasks, whose ideal heuristic would require significant domain knowledge.

\begin{table*}[!htb]
    \caption{Comparison of Contributions}
    \small
    \centering
    \begin{tabular}{lccccc}
        \toprule
        \textbf{Paper Name} & \textbf{Learning approach} & \textbf{Model type} & \textbf{Learned heuristic} & \textbf{Search type} & \textbf{Node expansion}\\
        \midrule
        SAIL & Interactive & Fully connected & Cost-to-go & Greedy search & All\\
        Neural A* & Supervised & U-Net & Guidance map & A* & Guidance map-based\\
        TransPath & Supervised & Transformer & Path probability & Focal search & $h_{FOCAL}$\\
        Discrepancy Focal Search & Supervised & Fully connected & Discrepancy & Focal search & $h_{FOCAL}$\\
        SLOPE (Ours) & Supervised & CNN regressor & Distance to $p*$ & Partial Expansion & Pruning\\
        \bottomrule
    \end{tabular}
    \label{tab:juxta}
\end{table*}
Multiple works have addressed learning a cost-to-go heuristic with deep learning methods to empower graph search algorithms and minimize the search effort. The most prominent one of them is SAIL \citep{choudhury2018data}, which trains a heuristic by imitating a clairvoyant oracle and establishes a map dataset that many works, including ours, use. A different data-driven approach is presented in Neural A* \citep{yonetani2021path}, which uses a canonical differentiable A* search as a loss to train an encoder (U-Net) to output a ``guidance map'' in which the search is conducted. The guidance map is used to add more cost to suboptimal nodes, making the search avoid them. This approach is not without its downsides, being very specific and hard to generalize or use with other search algorithms, as well as the environment having only unit node costs and the encoder having to be tailored to high-dimensional space. In \citep{kirilenko2023transpath} the authors propose the TransPath method, which uses a supervised learning approach that teaches a transformer model two heuristics, one being a correction factor of the instance-independent heuristic, and the other being a probability score of a node belonging to the shortest path. The optimal path can be sought with Theta* algorithm proposed in \cite{daniel2010theta} and the nodes that are close to the optimal path are added to the path probability data, otherwise, their probability is set to 0. This produces relatively narrow singular tunnels that could be susceptible to bottlenecking. The path probability score is then used as $h_{FOCAL}$ in the focal search. The work done by \cite{araneda2021exploiting} enhances focal search with a learned discrepancy heuristic, called Discrepancy Focal Search, similar to an optimal path probability. Their used metric, called discrepancy, is derived from the probability of a node's subpath being a part of an optimal path and is used as a heuristic in focal search.
Besides these, some other works treat different aspects but are still learning cost-to-go value are Neuro-algorithmic Policy (NAP) \citep{vlastelica2021neuroalgorithmic}, that is performing the planning on raw image inputs, \citet{groshev2018learning}, that used initial heuristics to solve problems and bootstrap the learning for solving more complicated problems, PHS \citep{ajanovic2023value}, that systematically explore the problem using Prolonged Heuristic Search,  Q* \citep{agostinelli2021search}, that learns one step cost together with cost-to-go and PHIL \citep{pandy2022learning}, that uses GNNs.

Related search algorithms include different partial expansion and bounded-cost search algorithms like focal search \citep{Pearl1982StudiesIS}, partial expansion A* \citep{yoshizumi2000partial} and recursive approaches like recursive best-first search \citep{KORF199341}. Focal search holds a focal list of open list nodes with $f$ values under a certain threshold and uses $h_{FOCAL}$ to order its elements for expansion. Partial Expansion A* (PEA*) is a modification of A* search that adds to the $\OPEN$ list only those nodes whose $f$ value is less than or equal to their parent nodes stored value plus some cutoff threshold C. If an expanded node has suboptimal child nodes, PEA* returns it to the $\OPEN$ list with the worst child's $f$ value. Recursive best-first search expands nodes with the best $f$ value that is also under a threshold and keeps track of previous second-best results. As soon as the best $f$ value is not monotonically decreasing, the algorithm recursively returns to the previous second-best node and continues the search from there.

In this work, we examine how a learned heuristic can direct the search effort to stay within an optimal region, effectively minimizing the number of expanded nodes and $\OPEN$ list size. Our heuristic is learned via a supervised approach and is not used in any way as a cost-to-go heuristic. We propose a simple metric to evaluate a node's ``optimality'', whose normalized value determines multiple gradual regions to which search can be limited. Having multiple gradual regions and the method of determining them distinguishes this work from previous ones, as we consider this approach to be more forgiving to different used cost-to-go heuristics and gives us the possibility to tune how strict we want our search to be based on the environment type. The region we consider ``optimal'' is wide enough to allow all possible shortest paths in situations when they are non-singular: something approaches with narrow optimal tunnels do not allow. Our $\textsc{SLOPE}$ search algorithms do not use the learned heuristic for any type of $\OPEN$ list ordering or deciding which node to expand- this we leave to a chosen cost-to-go heuristic with the goal of leveraging different learned information for a better search. Our search has built-in failsafe mechanisms that assure search completeness, even when region estimations have a bottleneck or are not connected with the goal.
For a more concise comparison, Table \ref{tab:juxta} presents key elements of the most relevant works and this paper.
To summarize, our contributions are as follows:
\begin{itemize}
    \item A simple approach to exploring a problem instance and generating the region containing all possible optimal paths, as well as neighboring regions that are $k$ steps away from the nearest optimal path.
    \item Training ML model based on the generated dataset to evaluate a node's optimality score that is based on the normalized distance to the optimal path.
    \item Two variants of a pruning search algorithm, that utilize the trained model and discard nodes whose score is under a certain threshold, effectively lowering both the number of expanded nodes as well as the $\OPEN$ list size.
    \item Exhaustive evaluation is done on multiple datasets, where we showcase the advantages of using our policy, both with a standard Euclidean heuristic as well as in combination with learned cost-to-go heuristics.
\end{itemize}

\section{Problem Statement}

\subsection{Search-Based Planing}

\begin{figure*}[t]
    \centering
    \includegraphics[width=\textwidth]{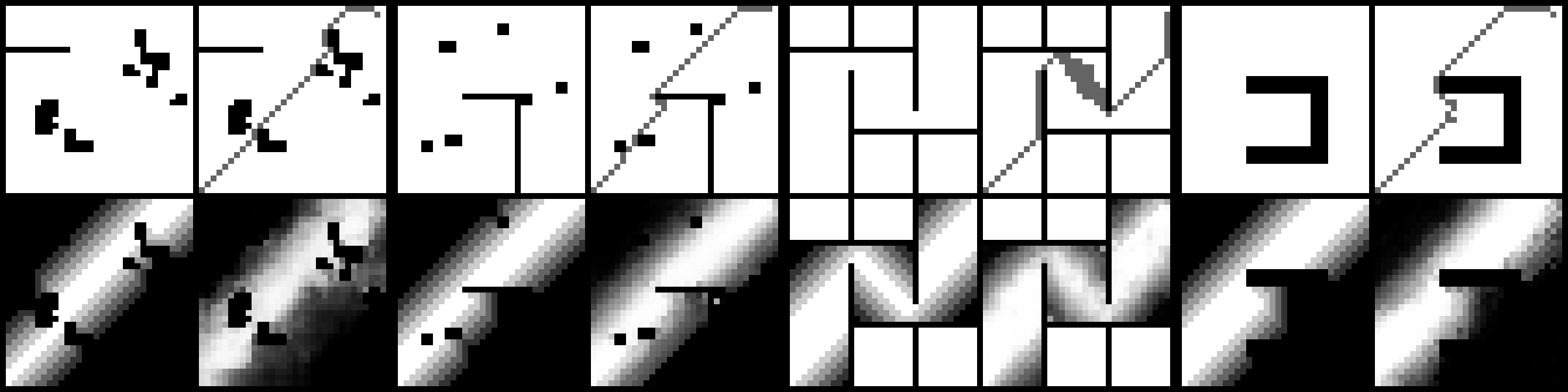}
    \caption{Visualizations of the steps of our method on several map types- forest, bugtrap+forest, maze and single bugtrap. The visualization elements are: map (upper left square), expanded nodes of our algorithm (upper right), dataset ground truth optimal areas (lower left) and model estimation of optimal areas (lower right).}
    \label{fig:banner_examples}
\end{figure*}

We consider the problem of planning based on the graph search, as shown in the Algorithm \ref{alg:search}.
At each iteration, successor nodes are generated in the function $\texttt{Expand}$ by expanding the \textit{current node} $\node$ using a transition model $\model$ to all reachable neighboring states. Each reachable collision-free state is represented with one \textit{child node}. All collision-free \textit{child nodes} $\childnodes$ are processed and are added to the $\OPEN$ list, unless they are already there. If the \textit{child node} is already in the $\OPEN$ list, and the new \textit{child node} has a lower cost, the parent of that node is updated, otherwise, it is ignored. From the $\OPEN$ list, at every iteration, $\texttt{Select}$ function extracts the node with the lowest cost, which is then chosen to be the next \textit{current node}, and the procedure is repeated until the goal is reached, the whole graph is explored or the computation time limit for planning is reached. At the end of the planning, if successful, the path is reconstructed starting from the goal $\nodeG$ iteratively towards the $\nodeI$.

\begin{algorithm}[t]
    \DontPrintSemicolon
    \fontsize{8pt}{9pt}\selectfont
    \SetKwData{n}{$n$}
    \SetKwFunction{Search}{Search}
    \SetKwFunction{Expand}{Expand}
    \SetKwFunction{Col}{ColCheck}
    \SetKwFunction{Select}{Select}
    \SetKwFunction{Children}{Children}
    \SetKwFunction{GetPath}{GetPath}
    \SetKwFunction{Update}{Update}
    \SetKwInOut{Input}{input}\SetKwInOut{Output}{output}
    \Input{$\nodeI$, $\nodeG$, $\OBST$, $\model$, $h(\cdot)$}
\BlankLine
\Begin{
    $\OPEN \gets \node \gets  \nodeI$\tcp*[r]{initialization}
    $\CLOSED \gets \varnothing$\;    
    \BlankLine		
    \While{$n \neq  \nodeG$ {\bf and} $ \OPEN \neq \varnothing$ {\bf and}  $ \CLOSED.size()\leq N_\mathrm{max} $}{
        $n \gets \Select(\OPEN)$\;
        $\OPEN \gets \OPEN \setminus n$\;	
        $\CLOSED \gets \CLOSED \; \cup \; n$\;
        $( \childnodes, \childcolnodes)\gets $  \Expand{$n, \OBST, \model, h(\node)$}\; 	
        $\CLOSED \gets \CLOSED \; \cup \;\childcolnodes$\;		
        \ForEach{$\node' \in \childnodes$}{
            $\OPEN \gets \Update (\OPEN , \node')$\;		
        }
    }
    \Return{$\GetPath(\nodeG)$}\tcp*[r]{reconstruct the path}
}
\caption{$\Search$: search-based planning. \label{alg:search}}
\end{algorithm}

\subsection{Learning Optimal Search}

In this work, we focus on the Imitation Learning \cite{osa2018algorithmic} approach for learning optimal search. In contrast to previous works in the field, we do not learn the heuristic function to estimate the exact cost-to-go. Rather we are learning a state-based function $d(n)$ that represents the distance from the closest optimal path.

The problem of Learning Optimal Search can be split into four subproblems: 
\begin{itemize}
    \item Curriculum design of example scenarios and scenario-variations;
    \item Scenario Exploration for oracle data generation;
    \item Supervised Learning to imitate the oracle;
    \item Using learned function in the search;
\end{itemize}
    
We assume having the fixed set of training problem instances from a single domain - curriculum similar to \citep{bhardwaj2017learning}. The goal is to efficiently explore them and learn reusable knowledge that can be utilized to have more efficient search in novel instances from the same domain.



\section{ Proposed method}

The presented approach for learning partial expansion explores a known model $\model$ of the system to generate dataset $\Dataset$ of exact state-distance $(\node, d(\node))$ data points. The dataset is used for supervised learning of the model $d_{ML}$. The learned model $d_{ML}$ is then used as in the heuristic search to prune unfavorable nodes, based on their estimated distance from the optimal path.
The dataset $\Dataset$ consists of data points that carry information about the scenario (obstacles $\OBST$, initial state $\nodeI$, and goal state $\nodeG$) and current state $n$ together with the corresponding distance label (i.e.\ shortest path from the current state to any state on the optimal path $d(n)$).
\subsection{Dataset generation}


To train the appropriate model to estimate the distance $d_{ML}(n)$ of a state to a potential optimal path, we need to generate an adequate dataset. 

To that end, we design a data pipeline that takes in different maps, generates the optimal cost-to-go values $h^*(n)$ for all reachable states within the map, and uses these values to find the optimal path region. Then, the shortest distance is computed for the nodes up to $m$  steps from the optimal region and the rest of the map is rated as far-from-optimal.

For generating the optimal cost-to-go we employ a prolonged Dijkstra-like algorithm backward; starting from the goal position, we iteratively move a frontier until reaching the start node. We prolong this search after reaching the start to generate enough samples for multiple regions.

The next part of dataset processing is extracting from the map the ``optimal path region'', which consists of all the nodes that belong to some optimal path. This is done by running a Dijkstra-like search algorithm from the start that computes the optimal cost-to-come, or distance from start, $g^*(n)$. The states that belong to the ``optimal path region'' all have $g^*(n)+h^*(n)$ constant.
The ``optimal path region'' is then extended in a similar frontier search fashion to $m$ ``neighboring regions''- which consist of points that are $1,2,...,m$ steps away from the nearest node in the optimal region. The rest of the map is considered far from optimal. These regions of nodes with the same distance from an optimal path we refer to as ``optimality regions''. For easier learning and usage, we normalize $d(n)$ by the maximum value $m$, with points in the optimal region having the value $1$, and far from optimal points having the value $0$. We refer to this value as a node's ``optimality rating''.

It is worth noting that we calculate the proposed metric from cost-to-go data via described transformations, and as such should be as difficult to learn as a cost-to-go heuristic. Conversely, the method with which we define our regions for dataset creation is based on Dijkstra search, and as such scales as well as Dijkstra. Overall, we can conclude the proposed method can scale to more complex problems as well as other supervised cost-to-go approaches.\\

\subsection{SLOPE algorithm}
To properly utilize the models trained on the described dataset, we develop two algorithms: pruning search $\textsc{SLOPE}$, described in the Algorithm \ref{alg:prune-search} and recursive prune search $\textsc{SLOPEr}$, defined as Algorithm \ref{alg:recursive-search}; both based on greedy best-first search. We have chosen greedy over A* search because not only does it outperform A* in these specific tasks (as shown in the comparison table of \citet{bhardwaj2017learning}), but it's otherwise desirable feature of completeness is not of paramount importance to our approach. By pruning, we consider the act of evaluating the node's optimality rating using $d_{ML}$, and based on a threshold $\tau$ we either add it to the $\OPEN$ list or discard it (or in the case of the first algorithm, add it to the backup $\OPEN$ list). Unlike other approaches that give a probabilistic rating of a node belonging to the shortest path (rather binary categorization), our model is trained on data with a larger range of values representing path optimality. This allows the following search to explore a wider region (up to a point based on the set threshold) and enables better collaboration with the heuristic. We generate a wide optimal path region, made up of all possible non-singular shortest paths, which allows more space for the chosen heuristic to navigate, enabling plug-and-play ease of use of cost-to-go heuristics.
Pruning search, hereafter casually denoted as $\textsc{SLOPE}$, takes in a fixed threshold value, ranging from 0 to 1, and either adds a node to the $\OPEN$ list, if its rating is higher than $\tau$, or adds it to the backup $\OPEN$ list, which becomes the active $\OPEN$ list if all of the optimal nodes are expanded but the goal is not yet reached. In this situation, we also scale down the threshold to ensure the continuation of the search effort (in our implementation we halve $\tau$ every time we run out of nodes). With this threshold scaling we assure search completeness, as our search becomes classic greedy search when $\tau$ approaches $0$. The described algorithm keeps search efforts close-to-optimal, which is an improvement in itself, but it does not fully optimize memory consumption, considering unfavored nodes are still stored in a backup list. The advantage of this approach, however, is minimizing $\OPEN$ list size, which speeds up real-time operations with the list, such as sorting elements for node selection.\\
In our testing we found determining $\tau$ to be a map-based task, considering the model has more difficulty learning the optimality ratings of certain map types than others (for alternating gaps we can safely use a $\tau$ of $0.9$, while for bugtrap+forest maps we found $0.57$ to be the optimal threshold\footnote{For visual examples of mentioned map types refer to Table \ref{tab:normal-final}}). Because of this, we introduce a recursive variant of our algorithm, the recursive pruning search denoted by $\textsc{SLOPEr}$. Unlike the first algorithm, this one does not save nonoptimal nodes to a backup list. Instead, it 
starts with a high value of the threshold (we set it at $0.9$), and each time the search runs out of nodes in the $\OPEN$ list, it recursively calls again itself with a lowered $\tau$ (we decrement our threshold by $0.1$ each call). In this way, the threshold is recursively custom-tailored to each specific map instance, unlike the first approach that uses one threshold for all maps in a test set, conversely achieving better results, leveraging the notion that search with the highest threshold value will give the most optimal results. As we will see in the Experiments section, this proved to be true as this algorithm achieved overall the best results, however with slightly increased computational overhead cost stemming from its recursive nature. Unlike the first algorithm, $\textsc{SLOPEr}$ does introduce memory consumption improvements, as unfavored nodes are discarded, and it's best-case scenario is an improvement in both speed and memory costs. The worst-case scenario, however, does suffer from higher computational overhead. This approach does depend more on the quality of the learnt heuristic than the $\textsc{SLOPE}$ approach.

The most noticeable difference between the two approaches is the existence of a backup list in $\textsc{SLOPE}$, which in turn enables different effects of thresholding to be utilized, even though both approaches adapt the threshold when the $\OPEN$ list is empty. Proposed algorithms leverage the advantages of the learned heuristic in different fashions, achieving diverse results due to search space being uniquely limited by their respective thresholding approaches.

\begin{algorithm}[t]
    \DontPrintSemicolon
    \fontsize{8pt}{9pt}\selectfont
    \SetKwData{n}{$n$}
    \SetKwFunction{PruningSearch}{SLOPE}
    \SetKwFunction{Expand}{Expand}
    \SetKwFunction{Col}{ColCheck}
    \SetKwFunction{Select}{Select}
    \SetKwFunction{Children}{Children}
    \SetKwFunction{GetPath}{GetPath}
    \SetKwFunction{Update}{Update}
    \SetKwFunction{CheckChild}{CheckChild}
    \SetKwInOut{Input}{input}\SetKwInOut{Output}{output}
    \Input{$\nodeI$, $\nodeG$, $\OBST$, $\model$, $d$, $\tau$}
\BlankLine
\Begin{
    $\OPEN \gets  \nodeI$\tcp*[r]{initialization}
    $\CLOSED \gets \varnothing$\;
    $\OPEN' \gets \varnothing$
    \BlankLine		
    \While{$n \neq  \nodeG$ {\bf and} $ \OPEN \neq \varnothing$ }{
        $n \gets \Select(\OPEN)$\;
        $\OPEN \gets \OPEN \setminus n$\;	
        $\CLOSED \gets \CLOSED \; \cup \; n$\;
        $( \childnodes, \childcolnodes)\gets $  \Expand{$n, \OBST, \model, h(\node)$}\; 	
        $\CLOSED \gets \CLOSED \; \cup \;\childcolnodes$\;		

        \ForEach{$\node' \in \childnodes$}{
            \If{\CheckChild{$\node'$} {\bf and} $\node' \notin \CLOSED$}{
                \If{$d(\node') > \tau$}{
                    $\OPEN \gets \OPEN \; \cup \; \node'$\;
                }
                \Else{
                    $\OPEN' \gets \OPEN' \; \cup \;\node'$\;
                }
                }	
            }
        \If{$n \neq  \nodeG$ {\bf and} $ \OPEN = \varnothing$ }{
            $\OPEN \gets \OPEN'$\;
            $\OPEN' \gets \varnothing$\;
            $\tau \gets \tau / 2$\;
        }
    }
    \Return{$\GetPath(\nodeG)$}\tcp*[r]{reconstruct the path}
}
\caption{$\PruningSearch$: pruned search-based planning. \label{alg:prune-search}}
\end{algorithm}

\begin{algorithm}[t]
    \DontPrintSemicolon
    \fontsize{8pt}{9pt}\selectfont
    \SetKwData{n}{$n$}
    \SetKwFunction{RecursivePruningSearch}{SLOPEr}
    \SetKwFunction{Expand}{Expand}
    \SetKwFunction{Col}{ColCheck}
    \SetKwFunction{Select}{Select}
    \SetKwFunction{Children}{Children}
    \SetKwFunction{GetPath}{GetPath}
    \SetKwFunction{Update}{Update}
    \SetKwFunction{CheckChild}{CheckChild}
    \SetKwInOut{Input}{input}\SetKwInOut{Output}{output}
    \Input{$\nodeI$, $\nodeG$, $\OBST$, $\model$, $d$, $\tau$}
\BlankLine
\Begin{
    $\OPEN \gets  \nodeI$\tcp*[r]{initialization}
    $\CLOSED \gets \varnothing$\;
    \BlankLine		
    \While{$n \neq  \nodeG$ {\bf and} $ \OPEN \neq \varnothing$ }{
        $n \gets \Select(\OPEN)$\;
        $\OPEN \gets \OPEN \setminus n$\;	
        $\CLOSED \gets \CLOSED \; \cup \; n$\;
        $( \childnodes, \childcolnodes)\gets $  \Expand{$n, \OBST, \model, h(\node)$}\; 	
        $\CLOSED \gets \CLOSED \; \cup \;\childcolnodes$\;		

        \ForEach{$\node' \in \childnodes$}{
            \If{\CheckChild{$\node'$} {\bf and} $\node' \notin \CLOSED$ {\bf and} $d(\node') > \tau$}{
                $\OPEN \gets \OPEN \; \cup \; \node'$\;
                }	
            }
        
    }
    \If{$n \neq  \nodeG$ {\bf and} $ \OPEN = \varnothing$ }{
        \Return{\RecursivePruningSearch{$\nodeI$, $\nodeG$, $\OBST$, $\model$, $\tau-\epsilon$}
        }}
    \Return{$\GetPath(\nodeG)$}\tcp*[r]{reconstruct the path}
}
\caption{$\RecursivePruningSearch$: recursive calling of pruned search-based planning. \label{alg:recursive-search}}
\end{algorithm}


\section{Experiments}
\subsection{Implementation details}

\begin{figure}[t]
    \centering
    \includegraphics[scale=0.3]{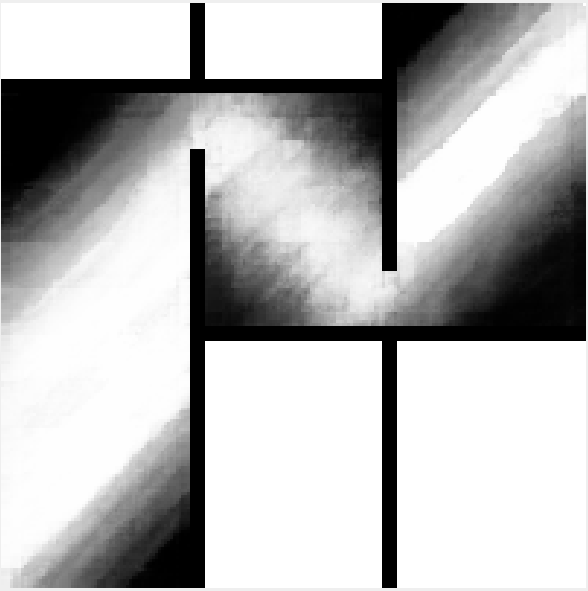}
    \caption{Learned optimality regions of a fully-sized 200x200 map. This model serves as a proof of concept, and as such was trained for 10 epochs on 100 maps, noticeably lower than our downscaled experiments.}
    \label{fig:fullsized}
\end{figure}

In this work, we use a collection of different instances from 8 grid domains, containing different obstacle types and map topologies and 8-connected cells, as used in SaIL \citep{bhardwaj2017learning}. We train and evaluate our ML models on 8 types of datasets. For simplicity and lowering computational costs, we rescale the original 200x200 maps to 32x32 environments (as in \citep{yonetani2021path}), as well as fix the starting point to the lower left corner and the goal to the upper right one. We use 320 maps for training, 80 maps for validation (80/20 split) and 100 test maps. For every reachable point of every map, we create a dedicated training sample by drawing in the point on the map and its optimality rating being the output, thus creating a dataset of hundreds of thousands of samples from just 400 maps. Considering different maps can have differently unbalanced region data (for instance, open maps such as single bugtrap or random forest have much more nonoptimal nodes than optimal ones), we downscale the most represented region down to the second most represented one per map with random sampling. It is worth noting that experiments with heavy downsampling of the regions and adding transformations to the input image were successful, proving that the model is able to learn multi-positional problems with a smaller dataset. Preliminary testing of learning the proposed heuristic on full-sized environments has been successful, proving the approach scales to larger environments. An example of this is shown on Figure \ref{fig:fullsized}. The ML model is based on 3 convolutional-maxpool blocks with 3 fully connected layers for regressing the convoluted image to the output ``optimality'' value. We train using Adam \citep{kingma2014adam} for 45 epochs with step learning rate reduction. Something to note: the focus of the paper was on developing the methodology, leaving the model architecture and training procedure room to improve.

\subsection{Metrics and baselines}
To evaluate and analyze the performance of our method, we use metrics based on three search result parameters: the number of expanded nodes, the length of the reconstructed path and the size of the $\OPEN$ list. Considering our data generation pipeline, we can extract the global minimal length of the optimal path. Using this, we define our three metrics as: the relative error of the number of expanded nodes w.r.t. the global minimum, the relative error of the length of the reconstructed path w.r.t. the global minimum and the size of the unused $\OPEN$ list normalized by the hypothetical full explorable space (set at 32x32=1024).

We compare our two main search methods, pruning with a preset fixed threshold $\textsc{SLOPE}$ and recursive pruning $\textsc{SLOPEr}$, both using Euclidean distance for the heuristic cost-to-go estimate. The $\textsc{SLOPEr}$ is expected better results. However, as this method is computationally more difficult, it is also beneficial to compare $h_{ML}$ and the pruning method $\textsc{SLOPE}$ with a fixed threshold. The optimal threshold value is manually obtained for specific dataset types, while not dipping below $0.45$ to preserve the point of this method.
We compare our algorithms to two greedy search baselines, one with Euclidean heuristic search $\greedyEuc$ and one with trained optimal cost-to-go $h_{ML}$. 
We train models that estimate optimal cost-to-go values generated from the first step of our dataset generation. We find this to be a suitable comparison, seeing how this is essentially the same data as the one for our SLOPE models, just used differently. 
To see how both our described search algorithms are orthogonal to ones that learn cost-to-go, and to prove our method can work in conjunction with the heuristic-based search, $\textsc{SLOPE}$ and $\textsc{SLOPEr}$ are benchmarked using $h_{ML}$ instead of Euclidean heuristic.

\begin{table*}[!h]
\small
\centering
\caption{Comparison of the relative error of expanded nodes (upper row, \%), relative error of the length of the reconstructed paths (middle row, \%) and normalized open list sizes (lower row) on 8 datasets and 6 methods. We pay special attention to comparing our approach combined with $h_{ML}$ to standard $h_{ML}$ (up/down arrows).}
\begin{tabulary}{1.0\textwidth}{LC|*{6}{C}} \toprule
 & {\bf World Samples} &{\bf $h_{ML}$} &{$SLOPE$} &{\bf $SLOPE+h_{ML}$} &{\bf $SLOPEr$ } &{\bf $SLOPEr+h_{ML}$} &{\bf $\greedyEuc$}\\ \hline 

{alternating gaps} &
\multirow{2}{*}{\raisebox{-0.5\totalheight}{\setlength{\fboxsep}{0pt}
		\setlength{\fboxrule}{1pt}\fbox{\includegraphics[width=0.04\textwidth, height=6mm]{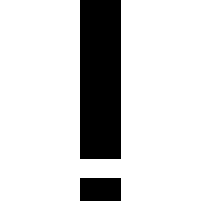}}} 
	\raisebox{-0.5\totalheight}{\setlength{\fboxsep}{0pt}
		\setlength{\fboxrule}{1pt}\fbox{\includegraphics[width=0.04\textwidth, height=6mm]{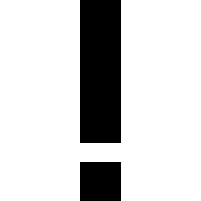}}} 
	\raisebox{-0.5\totalheight}{\setlength{\fboxsep}{0pt}
		\setlength{\fboxrule}{1pt}\fbox{\includegraphics[width=0.04\textwidth, height=6mm]{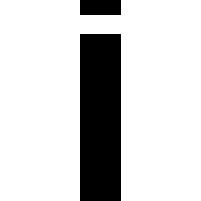}}}}
 &$0.845$ & \textbf{0.265} & $0.458 \downarrow$ &  \textbf{0.265} & $0.458 \downarrow$ & $340.135$\\
 
 & &$0.386$ & \textbf{0.072} & $0.193 \downarrow$ &  \textbf{0.072} & $0.193 \downarrow$ & $3.284$\\
 
 & &$0.118$ & \textbf{0.058} & $0.086 \downarrow$ &  \textbf{0.058} & $0.086 \downarrow$ & $0.122$\\ \hline

{shifting gaps} &
\multirow{2}{*}{\raisebox{-0.5\totalheight}{\setlength{\fboxsep}{0pt}
		\setlength{\fboxrule}{1pt}\fbox{\includegraphics[width=0.04\textwidth, height=6mm]{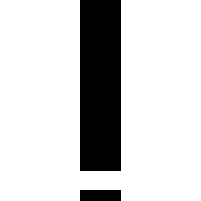}}} 
	\raisebox{-0.5\totalheight}{\setlength{\fboxsep}{0pt}
		\setlength{\fboxrule}{1pt}\fbox{\includegraphics[width=0.04\textwidth, height=6mm]{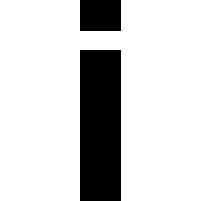}}} 
	\raisebox{-0.5\totalheight}{\setlength{\fboxsep}{0pt}
		\setlength{\fboxrule}{1pt}\fbox{\includegraphics[width=0.04\textwidth, height=6mm]{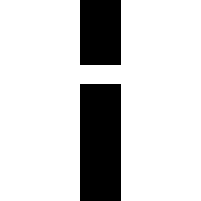}}}}
 &$0.360$ & \textbf{0} & $0.205 \downarrow$ &  \textbf{0} & $0.205 \downarrow$ & $149.871$\\
 
 & &$0.257$ & \textbf{0} & $0.051 \downarrow$ &  \textbf{0} & $0.051 \downarrow$ & $1.519$\\
 
 & &$0.119$ & \textbf{0.053} & $0.078 \downarrow$ &  \textbf{0.053} & $0.078 \downarrow$ & $0.110$\\ \hline

{single bugtrap} &
\multirow{2}{*}{\raisebox{-0.5\totalheight}{\setlength{\fboxsep}{0pt}
		\setlength{\fboxrule}{1pt}\fbox{\includegraphics[width=0.04\textwidth, height=6mm]{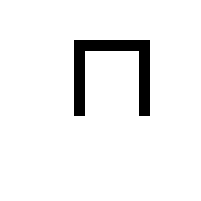}}} 
	\raisebox{-0.5\totalheight}{\setlength{\fboxsep}{0pt}
		\setlength{\fboxrule}{1pt}\fbox{\includegraphics[width=0.04\textwidth, height=6mm]{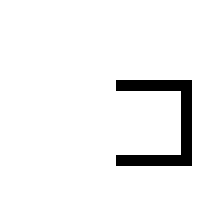}}} 
	\raisebox{-0.5\totalheight}{\setlength{\fboxsep}{0pt}
		\setlength{\fboxrule}{1pt}\fbox{\includegraphics[width=0.04\textwidth, height=6mm]{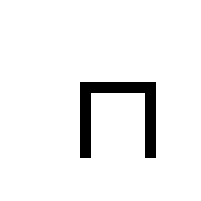}}}}
 &$2.619$ &$7.909$ &\textbf{2.515} $\downarrow$ &$3.915$ &$17.479 \uparrow$ & $56.587$\\
 
 &  &$1.218$ &$0.985$ &$1.270 \uparrow$ &\textbf{0.881} &$2.930 \uparrow$ & $3.993$\\
 
 & &$0.120$ &$0.083$ &$0.119 \downarrow$ &\textbf{0.070} &$0.087 \downarrow$ & $0.108$\\ \hline

{forest} &
\multirow{2}{*}{\raisebox{-0.5\totalheight}{\setlength{\fboxsep}{0pt}
		\setlength{\fboxrule}{1pt}\fbox{\includegraphics[width=0.04\textwidth, height=6mm]{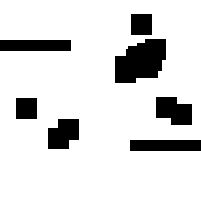}}} 
	\raisebox{-0.5\totalheight}{\setlength{\fboxsep}{0pt}
		\setlength{\fboxrule}{1pt}\fbox{\includegraphics[width=0.04\textwidth, height=6mm]{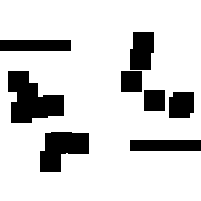}}} 
	\raisebox{-0.5\totalheight}{\setlength{\fboxsep}{0pt}
		\setlength{\fboxrule}{1pt}\fbox{\includegraphics[width=0.04\textwidth, height=6mm]{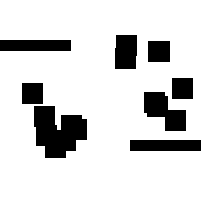}}}}
 &$13.054$ & $11.336$ & $15.030 \uparrow$ &  $10.277$ & $18.923 \uparrow$ & \textbf{5.410}\\
 
 & &$6.069$ & \textbf{1.889} & $6.355 \uparrow$ &  $2.519$ & $7.414 \downarrow$ &$2.920$\\
 
 & &$0.113$ & $0.101$ &$0.110 \downarrow$ &\textbf{0.092} & $0.098 \downarrow$ &$0.105$\\ \hline

{bugtrap+forest} &
\multirow{2}{*}{\raisebox{-0.5\totalheight}{\setlength{\fboxsep}{0pt}
		\setlength{\fboxrule}{1pt}\fbox{\includegraphics[width=0.04\textwidth, height=6mm]{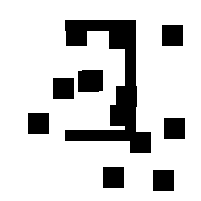}}} 
	\raisebox{-0.5\totalheight}{\setlength{\fboxsep}{0pt}
		\setlength{\fboxrule}{1pt}\fbox{\includegraphics[width=0.04\textwidth, height=6mm]{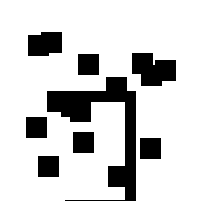}}} 
	\raisebox{-0.5\totalheight}{\setlength{\fboxsep}{0pt}
		\setlength{\fboxrule}{1pt}\fbox{\includegraphics[width=0.04\textwidth, height=6mm]{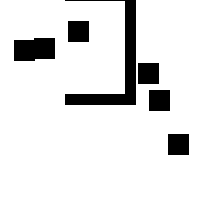}}}}
 &$23.323$ & $41.031$ & $32.977 \uparrow$ &  \textbf{13.998} & $24.593 \uparrow$ &$94.359$\\
 
 & &$9.273$ & $3.912$ & $9.349 \uparrow$ &  \textbf{2.261} & $8.536 \downarrow$ &$11.839$\\
 
 & &$0.122$ & $0.105$ & $0.122 \downarrow$ &  \textbf{0.085} & $0.092 \downarrow$ &$0.107$\\ \hline

{gaps+forest} &
\multirow{2}{*}{\raisebox{-0.5\totalheight}{\setlength{\fboxsep}{0pt}
		\setlength{\fboxrule}{1pt}\fbox{\includegraphics[width=0.04\textwidth, height=6mm]{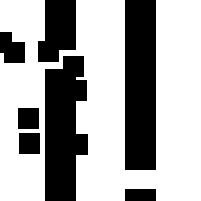}}} 
	\raisebox{-0.5\totalheight}{\setlength{\fboxsep}{0pt}
		\setlength{\fboxrule}{1pt}\fbox{\includegraphics[width=0.04\textwidth, height=6mm]{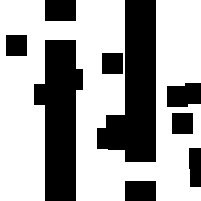}}} 
	\raisebox{-0.5\totalheight}{\setlength{\fboxsep}{0pt}
		\setlength{\fboxrule}{1pt}\fbox{\includegraphics[width=0.04\textwidth, height=6mm]{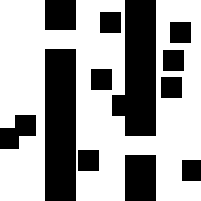}}}}
 &$26.733$ & $154.844$ & $49.641 \uparrow$ &  $101.068$ & \textbf{22.317} $\downarrow$ &$261.243$\\
 
 & &$6.061$ & $4.978$ & $5.132 \downarrow$ &  \textbf{3.009} & $3.220 \downarrow$ &$20.053$\\
 
 & &$0.183$ & $0.112$ & $0.180 \downarrow$ &  \textbf{0.076} & $0.142 \downarrow$ &$0.110$\\ \hline

{maze world} &
\multirow{2}{*}{\raisebox{-0.5\totalheight}{\setlength{\fboxsep}{0pt}
		\setlength{\fboxrule}{1pt}\fbox{\includegraphics[width=0.04\textwidth, height=6mm]{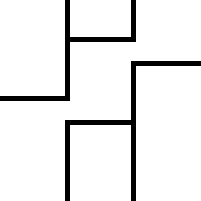}}} 
	\raisebox{-0.5\totalheight}{\setlength{\fboxsep}{0pt}
		\setlength{\fboxrule}{1pt}\fbox{\includegraphics[width=0.04\textwidth, height=6mm]{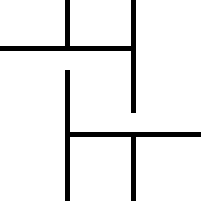}}} 
	\raisebox{-0.5\totalheight}{\setlength{\fboxsep}{0pt}
		\setlength{\fboxrule}{1pt}\fbox{\includegraphics[width=0.04\textwidth, height=6mm]{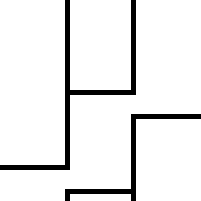}}}}
 &\textbf{1.615} & $22.739$ & \textbf{1.615} $\downarrow$ &  $16.783$ & $1.687 \uparrow$ &$44.393$\\
 
 & &\textbf{1.085} & $2.700$ & $1.109 \uparrow$ &  $1.929$ & $1.109 \uparrow$ &$6.317$\\
 
 & &$0.125$ & $0.099$ & $0.121 \downarrow$ &  \textbf{0.069} & $0.091 \downarrow$ &$0.106$\\ \hline

{multiple bugtraps} &
\multirow{2}{*}{\raisebox{-0.5\totalheight}{\setlength{\fboxsep}{0pt}
		\setlength{\fboxrule}{1pt}\fbox{\includegraphics[width=0.04\textwidth, height=6mm]{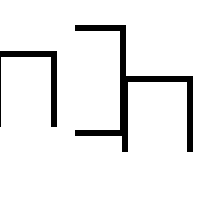}}} 
	\raisebox{-0.5\totalheight}{\setlength{\fboxsep}{0pt}
		\setlength{\fboxrule}{1pt}\fbox{\includegraphics[width=0.04\textwidth, height=6mm]{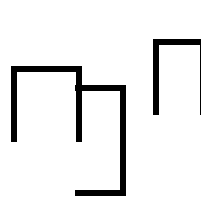}}} 
	\raisebox{-0.5\totalheight}{\setlength{\fboxsep}{0pt}
		\setlength{\fboxrule}{1pt}\fbox{\includegraphics[width=0.04\textwidth, height=6mm]{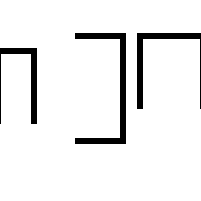}}}}
 &$130.085$ & $37.554$ & $70.075 \downarrow$ & \textbf{21.742} & $96.046 \downarrow$ &$205.722$\\
 
 & &$8.067$ & $4.435$ & $7.538 \downarrow$ &\textbf{2.229} & $4.205 \downarrow$ &$16.501$\\
 
 & &$0.118$ & $0.100$ & $0.109 \downarrow$ &  \textbf{0.071} & $0.077 \downarrow$ &$0.111$\\ \hline

\bottomrule
\end{tabulary}
\label{tab:normal-final}
\end{table*}
\begin{table*}[!htpb]
\small
\centering
\caption{Comparison of results obtained with trained pruning model and ground truth annotations, both standalone and in combination with ML heuristic. We can note ground truth experiments having the best results, further proving the methodology and capabilites of combining with ML heuristics.}
\begin{tabulary}{1.0\textwidth}{LC|*{5}{C}} \toprule
 & {\bf World Samples} &{\bf $h_{ML}$} &{\bf $SLOPE_{ML}$} &{\bf $SLOPE_{ML}+h_{ML}$} &{\bf $SLOPE_{GT}$ } &{\bf $SLOPE_{GT}+h_{ML}$}\\ \hline

{single bugtrap} &
\multirow{2}{*}{\raisebox{-0.5\totalheight}{\setlength{\fboxsep}{0pt}
		\setlength{\fboxrule}{1pt}\fbox{\includegraphics[width=0.04\textwidth, height=6mm]{figures/database_samples/bugtrap_world/1.png}}} 
	\raisebox{-0.5\totalheight}{\setlength{\fboxsep}{0pt}
		\setlength{\fboxrule}{1pt}\fbox{\includegraphics[width=0.04\textwidth, height=6mm]{figures/database_samples/bugtrap_world/2.png}}} 
	\raisebox{-0.5\totalheight}{\setlength{\fboxsep}{0pt}
		\setlength{\fboxrule}{1pt}\fbox{\includegraphics[width=0.04\textwidth, height=6mm]{figures/database_samples/bugtrap_world/3.png}}}}
 &$2.619 $ &$7.909$ &$2.515 \downarrow$ &\textbf{1.296} &$1.919 \downarrow$\\
 & &$1.218 $ &$0.985$ &$1.270 \uparrow$ &\textbf{0} &$1.218 \downarrow$\\
 & &$0.120$ &$0.083$ &$0.119 \downarrow$ &\textbf{0.054} &$0.119 \downarrow$\\ \hline

{forest} &
\multirow{2}{*}{\raisebox{-0.5\totalheight}{\setlength{\fboxsep}{0pt}
		\setlength{\fboxrule}{1pt}\fbox{\includegraphics[width=0.04\textwidth, height=6mm]{figures/database_samples/forest_world/1.png}}} 
	\raisebox{-0.5\totalheight}{\setlength{\fboxsep}{0pt}
		\setlength{\fboxrule}{1pt}\fbox{\includegraphics[width=0.04\textwidth, height=6mm]{figures/database_samples/forest_world/2.png}}} 
	\raisebox{-0.5\totalheight}{\setlength{\fboxsep}{0pt}
		\setlength{\fboxrule}{1pt}\fbox{\includegraphics[width=0.04\textwidth, height=6mm]{figures/database_samples/forest_world/3.png}}}}
 &$13.054$ & $11.336$ & $15.030 \uparrow$ &  \textbf{0.085} & $2.891 \downarrow$ \\
 & &$6.069$ & $1.889$ & $6.355 \uparrow$ &  \textbf{0} & $0.944 \downarrow$ \\
 & &$0.113$ & $0.101$ &$0.110 \downarrow$ &\textbf{0.047} & $0.054 \downarrow$\\ \hline

{multiple bugtraps} &
\multirow{2}{*}{\raisebox{-0.5\totalheight}{\setlength{\fboxsep}{0pt}
		\setlength{\fboxrule}{1pt}\fbox{\includegraphics[width=0.04\textwidth, height=6mm]{figures/database_samples/multiple_bugtrap_world/1.png}}} 
	\raisebox{-0.5\totalheight}{\setlength{\fboxsep}{0pt}
		\setlength{\fboxrule}{1pt}\fbox{\includegraphics[width=0.04\textwidth, height=6mm]{figures/database_samples/multiple_bugtrap_world/2.png}}} 
	\raisebox{-0.5\totalheight}{\setlength{\fboxsep}{0pt}
		\setlength{\fboxrule}{1pt}\fbox{\includegraphics[width=0.04\textwidth, height=6mm]{figures/database_samples/multiple_bugtrap_world/3.png}}}}
 &$130.085$ & $37.554 $ & $70.075 \downarrow$ &  \textbf{9.813} & $62.790 \downarrow$ \\
 & &$8.067 $ & $4.435 $ & $7.538 \downarrow$ &  \textbf{0} & $7.423 \downarrow$ \\
 & &$0.118 $ & $0.100 $ & $0.109 \downarrow$ &  \textbf{0.070} & $0.077 \downarrow$\\ \hline

{gaps+forest} &
\multirow{2}{*}{\raisebox{-0.5\totalheight}{\setlength{\fboxsep}{0pt}
		\setlength{\fboxrule}{1pt}\fbox{\includegraphics[width=0.04\textwidth, height=6mm]{figures/database_samples/gap_world/1.png}}} 
	\raisebox{-0.5\totalheight}{\setlength{\fboxsep}{0pt}
		\setlength{\fboxrule}{1pt}\fbox{\includegraphics[width=0.04\textwidth, height=6mm]{figures/database_samples/gap_world/2.png}}} 
	\raisebox{-0.5\totalheight}{\setlength{\fboxsep}{0pt}
		\setlength{\fboxrule}{1pt}\fbox{\includegraphics[width=0.04\textwidth, height=6mm]{figures/database_samples/gap_world/3.png}}}}
 &$26.733 $ & $154.844$ & $49.641 \uparrow$ &  $73.871$ & \textbf{22.106} $\downarrow$ \\
 & &$6.061 $ & $4.978$ & $5.132  \downarrow$ &  \textbf{0} & $4.992 \downarrow$ \\
 & &$0.183 $ & $0.112 $ & $0.180 \downarrow$ &  \textbf{0.058} & $0.170 \downarrow$\\ \hline

{maze world} &
\multirow{2}{*}{\raisebox{-0.5\totalheight}{\setlength{\fboxsep}{0pt}
		\setlength{\fboxrule}{1pt}\fbox{\includegraphics[width=0.04\textwidth, height=6mm]{figures/database_samples/maze_world/1.png}}} 
	\raisebox{-0.5\totalheight}{\setlength{\fboxsep}{0pt}
		\setlength{\fboxrule}{1pt}\fbox{\includegraphics[width=0.04\textwidth, height=6mm]{figures/database_samples/maze_world/2.png}}} 
	\raisebox{-0.5\totalheight}{\setlength{\fboxsep}{0pt}
		\setlength{\fboxrule}{1pt}\fbox{\includegraphics[width=0.04\textwidth, height=6mm]{figures/database_samples/maze_world/3.png}}}}
 &$1.615 $ & $22.739$ & $1.615  \downarrow$ &  $6.679$ & \textbf{1.567} $\downarrow$ \\
 & &$1.085$ & $2.700$ & $1.109 \uparrow$ &  \textbf{0} & $1.085$ $\downarrow$ \\
 & &$0.125$ & $0.099$ & $0.121 \downarrow$ &  \textbf{0.052} & $0.124 \downarrow$\\ \hline
 
\bottomrule
\end{tabulary}
\label{tab:hml_gt_comp}
\end{table*}

\subsection{Results}
The performance indicators of the aforementioned methods are shown in Table \ref{tab:normal-final}. A visual representation of optimality regions, learned estimations of optimality regions and expanded nodes during search for certain map types can be seen on Figure \ref{fig:banner_examples}.
Firstly, by comparing the individual methods $h_{ML}$, $\textsc{SLOPE}$ and $\textsc{SLOPEr}$, we notice that $\textsc{SLOPEr}$ performs better on average. This is to be expected, as the pruning threshold is recursively adjusted for each individual map, with the highest threshold (up to a resolution step) that enables reaching the goal giving a better result. While comparing the expanded node metrics, we can see the pruning method achieves better results on maps such as alternating gaps, forest and multiple bug traps, but also achieves drastically worse results on  gaps+forest and maze maps. In these cases, the worse results are partly due to using the Euclidean heuristic, with these maps having inherent bug trap-like obstacles trapping the agent, even though most of the expanded nodes are in the optimal region. A worthy mention is both SLOPE algorithms achieve perfect results on the shifting gaps dataset, mostly due to the simple and repetitive nature of the map types.

\begin{figure}[t]
    \centering
    \includegraphics[scale=0.3]{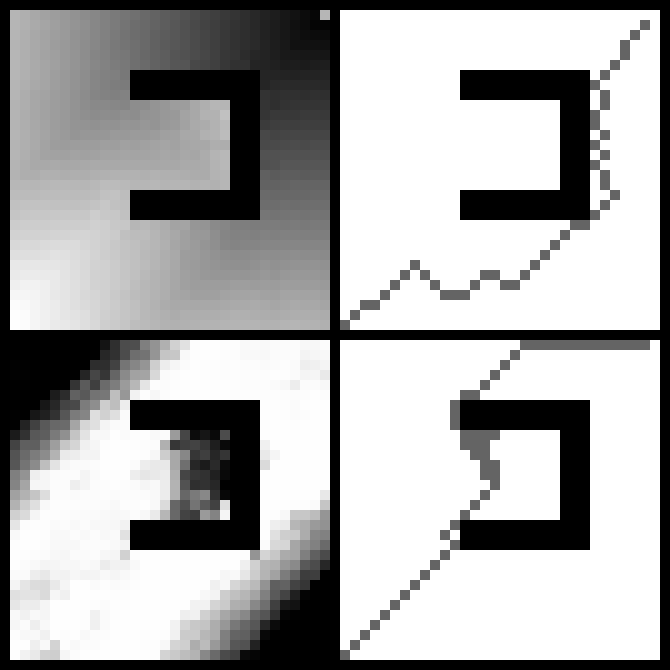}
    \caption{Visualization of the different approaches by the cost-to-go model (upper row) and $d_{ML}$ (lower row). The visualization of the cost-to-go model shows the model leaning towards the path under the bugtrap, while $d_{ML}$ is blocking that approach and taking the path above it. Combining the two, that are individually both sound, gives an in-between mix that is worse than both.}
    \label{fig:problem_compare}
\end{figure}

Further focusing on the comparison of the ML-powered SLOPE search with the base one, competitive node expansion metrics are achieved on maps such as alternating gaps, single bug trap, maze world and multiple bug traps, while also lowering $\OPEN$ list size for all. We notice the combined approach having the best or close to the best results of all methods on single bug trap, gaps+forest and maze world. The combined approach achieves worse results than individual methods on the forest dataset and in-between results on the rest of the datasets.
Thus we show the possibility of combining the two methods, weighing between optimizing the expanded nodes and minimizing the $\OPEN$ list, and in so, the computational cost.
With regards to the recursive pruning approach and its combination with the $h_{ML}$ for cost-to-go estimate, we note the same weighing effect on most maps, except for forest, single bug trap, bugtrap+forest, which were worse, and gaps+forest which gave better results than individual methods. The gaps+forest improvement comes from the mentioned problems with the Euclidean heuristic and them being solved using the ML heuristic. The problem with the bugtrap maps is illustrated in Figure \ref{fig:problem_compare}. The two models (cost-to-go and $d_{ML}$) lead the search in different directions, both optimal solutions, but significantly different, due to the existence of multi-modal solution. This in turn, when combining them, gives results that are suboptimal to both individual approaches.

Another possible reason for the combined results not being always better might be the inherent imperfection of the trained ML models, with certain estimation errors carrying great consequences in the pruning task. As an interesting experiment and a proof of concept of the fundamental method, we use the ground truth optimality region labels generated for the test set as a pruning heuristic and compare using it instead of the trained ML pruning, results of this are shown in Table \ref{tab:hml_gt_comp}. This way we remove any effective bias that results from the ML pruning model having faulty estimations. Here we see using the ground truth optimality regions achieves the best results in multiple map types, and its combination with ML heuristic achieves the best results in gaps+forest and maze world datasets, it being an improvement to Euclidean heuristic as explained earlier. These experiments also put in perspective results from Table \ref{tab:normal-final}, seeing how the ground truth experiments are the best-case scenario for the trained heuristic. This gives us insight into the level of involvement the Euclidean heuristic had in producing worse results on maze world and gaps+forest datasets. Both ground truth experiments beat out the ML heuristic approach on all datasets, which is what we aimed to prove.






\section{Conclusion}
In this paper, we presented a novel approach that learns a heuristic for node expansion. This aids heuristic search planners in solving grid-based pathfinding problems faster while also lowering the cost and resources needed for search efforts. We presented the advantages of this approach, mainly lowering expanded nodes and $\OPEN$ list size while also being open to combining with cost-to-go heuristics, effectively utilizing the best of both worlds. In experiments, we notice that the method is sensitive to bottlenecking of the optimal region. This occurs due to the inherent imperfection of the trained model's estimations. We solve this by adjusting the appropriate threshold, but another natural future direction would be enhancing the current model structure (which is a fairly simple CNN) and testing more advanced networks such as GNNs or Transformers, or different learning approaches such as Interactive and Reinforcement Learning. Furthermore, as our experiments were run on downscaled maps, that is something that should be revised in future works, as certain maps could yield better results with more navigation space allowed on full-sized maps. Considering our method doesn't assume a cost-to-go value, but rather a normalized score of a node, it could be adjusted to work with limited environment visibility, as it is easier to determine the distance from an optimal path than a cost-to-goal value without knowing the full map topology.





\bibliography{references}

\begin{thebibliography}{17}
\providecommand{\natexlab}[1]{#1}

\bibitem[{Agostinelli et~al.(2021)Agostinelli, Shmakov, McAleer, Fox, and
  Baldi}]{agostinelli2021search}
Agostinelli, F.; Shmakov, A.; McAleer, S.; Fox, R.; and Baldi, P. 2021.
\newblock A* search without expansions: Learning heuristic functions with deep
  q-networks.
\newblock \emph{arXiv preprint arXiv:2102.04518}.

\bibitem[{Ajanovic, Lacevic, and Kober(2023)}]{ajanovic2023value}
Ajanovic, Z.; Lacevic, B.; and Kober, J. 2023.
\newblock Value Function Learning via Prolonged Backward Heuristic Search.
\newblock In \emph{PRL Workshop Series $\{$$\backslash$textendash$\}$ Bridging
  the Gap Between AI Planning and Reinforcement Learning}.

\bibitem[{Araneda, Greco, and Baier(2021)}]{araneda2021exploiting}
Araneda, P.; Greco, M.; and Baier, J.~A. 2021.
\newblock Exploiting learned policies in focal search.
\newblock In \emph{Proceedings of the International Symposium on Combinatorial
  Search}, volume~12, 2--10.

\bibitem[{Bhardwaj, Choudhury, and Scherer(2017)}]{bhardwaj2017learning}
Bhardwaj, M.; Choudhury, S.; and Scherer, S. 2017.
\newblock Learning heuristic search via imitation.
\newblock In \emph{Conference on Robot Learning}, 271--280. PMLR.

\bibitem[{Choudhury et~al.(2018)Choudhury, Bhardwaj, Arora, Kapoor, Ranade,
  Scherer, and Dey}]{choudhury2018data}
Choudhury, S.; Bhardwaj, M.; Arora, S.; Kapoor, A.; Ranade, G.; Scherer, S.;
  and Dey, D. 2018.
\newblock Data-driven planning via imitation learning.
\newblock \emph{The International Journal of Robotics Research}, 37(13-14):
  1632--1672.

\bibitem[{Daniel et~al.(2010)Daniel, Nash, Koenig, and
  Felner}]{daniel2010theta}
Daniel, K.; Nash, A.; Koenig, S.; and Felner, A. 2010.
\newblock Theta*: Any-angle path planning on grids.
\newblock \emph{Journal of Artificial Intelligence Research}, 39: 533--579.

\bibitem[{Groshev et~al.(2018)Groshev, Tamar, Goldstein, Srivastava, and
  Abbeel}]{groshev2018learning}
Groshev, E.; Tamar, A.; Goldstein, M.; Srivastava, S.; and Abbeel, P. 2018.
\newblock Learning generalized reactive policies using deep neural networks.
\newblock In \emph{2018 AAAI Spring Symposium Series}.

\bibitem[{Hart, Nilsson, and Raphael(1968)}]{Hart1968AFB}
Hart, P.~E.; Nilsson, N.~J.; and Raphael, B. 1968.
\newblock A Formal Basis for the Heuristic Determination of Minimum Cost Paths.
\newblock \emph{IEEE Transactions on Systems Science and Cybernetics}, 4(2):
  100--107.

\bibitem[{Kingma and Ba(2014)}]{kingma2014adam}
Kingma, D.~P.; and Ba, J. 2014.
\newblock Adam: A method for stochastic optimization.
\newblock \emph{arXiv preprint arXiv:1412.6980}.

\bibitem[{Kirilenko et~al.(2023)Kirilenko, Andreychuk, Panov, and
  Yakovlev}]{kirilenko2023transpath}
Kirilenko, D.; Andreychuk, A.; Panov, A.; and Yakovlev, K. 2023.
\newblock Transpath: Learning heuristics for grid-based pathfinding via
  transformers.
\newblock In \emph{Proceedings of the AAAI Conference on Artificial
  Intelligence}, volume~37, 12436--12443.

\bibitem[{Korf(1993)}]{KORF199341}
Korf, R.~E. 1993.
\newblock Linear-space best-first search.
\newblock \emph{Artificial Intelligence}, 62(1): 41--78.

\bibitem[{Osa et~al.(2018)Osa, Pajarinen, Neumann, Bagnell, Abbeel, Peters
  et~al.}]{osa2018algorithmic}
Osa, T.; Pajarinen, J.; Neumann, G.; Bagnell, J.~A.; Abbeel, P.; Peters, J.;
  et~al. 2018.
\newblock An algorithmic perspective on imitation learning.
\newblock \emph{Foundations and Trends{\textregistered} in Robotics}, 7(1-2):
  1--179.

\bibitem[{P\'andy et~al.(2022)P\'andy, Qiu, Corso, Veli\v{c}kovi\'c, Ying,
  Leskovec, and Lio}]{pandy2022learning}
P\'andy, M.; Qiu, W.; Corso, G.; Veli\v{c}kovi\'c, P.; Ying, Z.; Leskovec, J.;
  and Lio, P. 2022.
\newblock Learning {{Graph Search Heuristics}}.

\bibitem[{Pearl and Kim(1982)}]{Pearl1982StudiesIS}
Pearl, J.; and Kim, J.~H. 1982.
\newblock Studies in Semi-Admissible Heuristics.
\newblock \emph{IEEE Transactions on Pattern Analysis and Machine
  Intelligence}, PAMI-4: 392--399.

\bibitem[{Vlastelica, Rolinek, and
  Martius(2021)}]{vlastelica2021neuroalgorithmic}
Vlastelica, M.; Rolinek, M.; and Martius, G. 2021.
\newblock Neuro-Algorithmic {{Policies Enable Fast Combinatorial
  Generalization}}.
\newblock In \emph{Proceedings of the 38th {{International Conference}} on
  {{Machine Learning}}}, 10575--10585. {PMLR}.

\bibitem[{Yonetani et~al.(2021)Yonetani, Taniai, Barekatain, Nishimura, and
  Kanezaki}]{yonetani2021path}
Yonetani, R.; Taniai, T.; Barekatain, M.; Nishimura, M.; and Kanezaki, A. 2021.
\newblock Path planning using neural A* search.
\newblock In \emph{International conference on machine learning}, 12029--12039.
  PMLR.

\bibitem[{Yoshizumi, Miura, and Ishida(2000)}]{yoshizumi2000partial}
Yoshizumi, T.; Miura, T.; and Ishida, T. 2000.
\newblock A* with Partial Expansion for Large Branching Factor Problems.
\newblock In \emph{AAAI/IAAI}, 923--929.

\end{thebibliography}

\end{document}